\documentclass[runningheads]{llncs}
\usepackage[table]{xcolor}
\definecolor{mycolor}{RGB}{230,230,255}
\usepackage[T1]{fontenc}
\usepackage{hyperref}
\usepackage{arydshln}
\usepackage{tablefootnote}
\usepackage{graphicx}
\usepackage{marvosym}
\usepackage{subcaption}
\usepackage{amsmath,amssymb,amsfonts}%
\usepackage{booktabs}
\usepackage{multirow}
\begin{document}
\title{PTransformer: A Prompt-based Multimodal Transformer Architecture For Medical Tabular Data}
\titlerunning{PTransformer}
%
\author{Yucheng Ruan\inst{1,2}$^\dagger$ \and
Xiang Lan\inst{1,2}$^\dagger$ \and
Daniel J. Tan\inst{2} \and
Hairil Rizal Abdullah\inst{3}\textsuperscript{\S}  \and
Mengling Feng\inst{1,2}\textsuperscript{\S (\Letter)} 
}
\authorrunning{Y. Ruan et al.}
%
\institute{Saw Swee Hock School of Public Health, National University of Singapore, Singapore \and
Institute of Data Science, National University of Singapore, Singapore \and
Department of Anaesthesiology, Singapore General Hospital, Singapore \\
\email{ephfm@nus.edu.sg} \\
$\dagger$ Equal contributions \\
$\S$ Co-senior authors
}
\maketitle              
\begin{abstract}
Medical tabular data, abundant in Electronic Health Records (EHRs), is a valuable resource for diverse medical tasks such as risk prediction. While deep learning approaches, particularly transformer-based models, have shown remarkable performance in tabular data prediction, there are still problems remaining for existing work to be effectively adapted into medical domain, such as ignoring unstructured free-texts and underutilizing the textual information in structured data. To address these issues, we propose PTransformer, a \underline{P}rompt-based multimodal \underline{Transformer} architecture designed specifically for medical tabular data. This framework consists of two critical components: a tabular cell embedding generator and a tabular transformer. The former efficiently encodes diverse modalities from both structured and unstructured tabular data into a harmonized language semantic space with the help of pre-trained sentence encoder and medical prompts. The latter integrates cell representations to generate patient embeddings for various medical tasks. In comprehensive experiments on two real-world datasets for three medical tasks, PTransformer demonstrated the improvements with 10.9\%/11.0\% on RMSE/MAE, 0.5\%/2.2\% on RMSE/MAE, and 1.6\%/0.8\% on BACC/AUROC compared to state-of-the-art (SOTA) baselines in predictability. 

\end{abstract}
\newpage
\section{Introduction}
In recent years, electronic health records (EHRs) have rapidly expanded in medical institutions, causing a significant increase in the amount and complexity of medical data. Among the different types of medical data, tabular data is particularly valuable because it includes diverse patient details, clinical observations, and diagnostic outcomes \cite{przystalski2023medical}. Analyzing the medical tabular data can reveal important patterns and insights that are useful for healthcare providers, researchers, and policymakers. Therefore, a comprehensive understanding of the complexities in medical tabular data is crucial to fully harness its potential and improve patient outcomes through data-driven insights.

Existing machine learning research has showed the predictive superiority of tree-based approaches \cite{grinsztajn2022tree}. However, the predominant focus of these approaches has been on structured medical data (e.g. categorical, numerical, and binary data types) \cite{gao2022prediction}. 
With the rise of deep learning, researchers have developed a wide range of frameworks, with transformer-based architectures notably standing out, while still primarily focusing on structured data \cite{arik2021tabnet,huang2020tabtransformer,gorishniy2021revisiting}. 
Despite of the prevalent structured medical tabular data, we believe that unstructured clinical free-text data, such as clinical notes, diagnostic tests, and preoperative diagnoses, contains valuable clinical information, which can be incorporated to enhance overall performance on medial tabular data modeling.

In addition, there is an underutilization of textual information within structured data in existing research. For example, categorical features such as "diagnosis" and "prescribed medication" often involve textual descriptions of a patient's condition or treatment context. These descriptions are usually encoded numerically during modeling, ignoring the semantic meanings or hierarchical relationships between different categories.

\begin{figure*}[!h]
\vspace{-0.5cm}
\centering
\begin{center}
    \includegraphics[scale = 0.3]{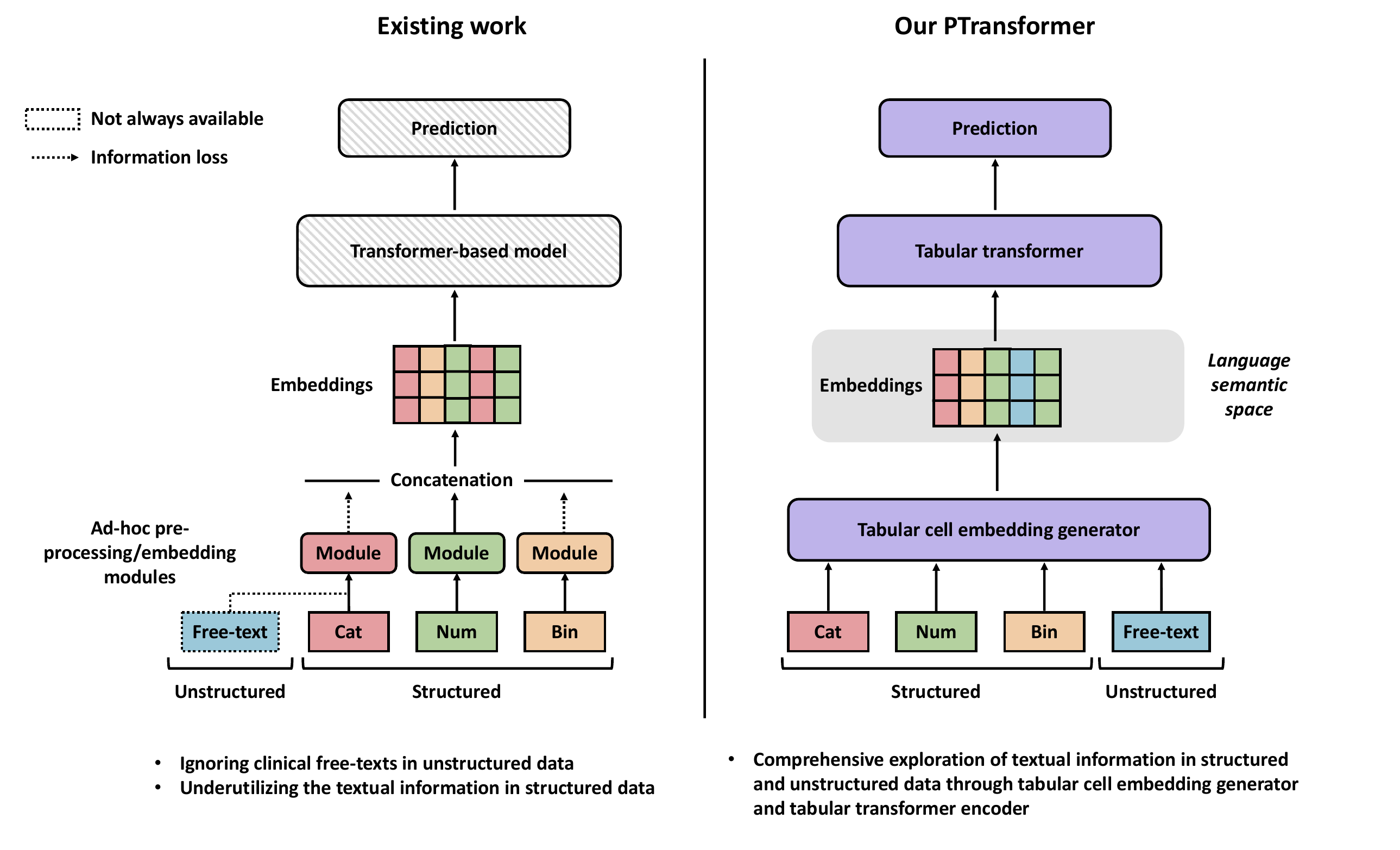}
\end{center}

\caption{The comparison between existing work and our proposed model. Main modules of the proposed framework: (1) Tabular cell embedding generator, (2) Tabular transformer, (3) Prediction head. 
}

  \label{fig:framework_comparision}
  \vspace{-0.5cm}
\end{figure*}

To address the aforementioned issues, we propose PTransformer, a novel \underline{P}rompt-based multimodal tabular \underline{Transformer} architecture for medical tabular data, which consists of two major components: tabular cell embedding generator and tabular transformer. 
In tabular cell embedding generator, all cell values from various modalities (continuous, categorical, binary, and free-texts) are processed as texts such that the frozen pre-trained sentence encoder is able to explore the textual information and extract the cell embeddings in the harmonized language semantic space for feature representation. 
In order to produce optimal cell embeddings with the pre-trained encoder, we introduce the prompts construction module that transform cell values into natural sentences with our designed medical prompts.
The medical prompts helps to extract most relevant information from the pre-trained sentence encoder in the medical context, thus incorporating additional clinical information into the model. 
In the tabular transformer, all cell embeddings from different modalities in the harmonized language latent space are fused and integrated to generate high-level patient embedding for making predictions with a shallow network.
Our model serves as the backbone of modeling different modalities in medical tabular data and can be easily extended for various medical tasks. 
Figure \ref{fig:framework_comparision} provides a detailed comparison between the existing methodologies and our proposed PTransformer.

\section{Related Work}

\subsection{Medical Tabular Prediction}
In medical tabular prediction, machine learning approaches have been extensively investigated in various tasks. Nistal-Nuño \cite{nistal2022developing} established Bayesian Network, Naïve Bayes network, and XGBoost model to assess the risk of mortality in the ICU using physiological measurements, demographic and diagnoses features. Additionally, Gao et al. \cite{gao2022prediction} explored an ensemble method based on gradient boosting decision tree algorithms for early prediction of acute kidney injury occurrence.

With the advent of deep learning, various frameworks tailored for the analysis of medical tabular data have emerged. Chen et al. \cite{chen2021automatic} presented a Multilayer Perceptron (MLP) to construct a surgery duration prediction system using several demographics and clinical features. George et al. \cite{george2021deep} developed a feed-forward neural network for predicting 3-month mortality in patients requiring 7 days of mechanical ventilation, utilizing demographic, physiologic, and clinical data. 

\subsection{Transformer-based Models for Tabular Data Modeling}
Transformer \cite{vaswani2017attention} is a prominent deep learning
model, which revolutionized NLP on a wide range of language tasks. 
It gains the popularity in modeling tabular data, and recent transformer-based studies show better performance than other machine learning models. 
TabNet \cite{arik2021tabnet} is a pioneering transformer-based model using sequential attention for tabular data prediction. 
TabTransformer \cite{huang2020tabtransformer} uses self-attention transformers to convert categorical features into contextual embeddings, improving predictive performance. FT-Transformer \cite{gorishniy2021revisiting} introduces the Feature Tokenizer to transform both categorical and numerical features into embeddings, which are well explored through multiple Transformer layers for accurate predictions.

Despite the advancements in transformer-based models, adapting these approaches to the medical tabular domain comprehensively remains challenging. These models mainly focus on structured tabular data, often overlooking valuable information in free-texts within medical tabular data.
Additionally, existing work underutilizes textual information within structured data.
Wang et al. \cite{wang2022transtab} address this limitation by adapting word embeddings to integrate textual information from both structured data and unstructured texts.
However, the challenge persists in fully capturing textual information in embeddings, especially when there is a lack of extensive training data typically required for language model pre-training. Our objective is to develop a universal embedding module with a transformer-based architecture capable of representing all patient characteristics in a harmonized language space within medical tabular data, effectively exploring textual information in both structured data and unstructured free-texts.

\section{Methodology}\label{sec3}

\subsection{Overview}
In this framework, we first develop a tabular embedding generator, which incorporates a pre-trained sentence encoder with a specifically designed prompt construction module to extract contextual cell embeddings from medical tabular data. Next, the generated cell embeddings are fed into a tabular transformer, which helps produce informative patient embeddings for different medical tasks. The architecture overview is depicted in Figure \ref{fig:framework_comparision}.

\subsection{Tabular Cell Embedding Generator}

\begin{figure*}[!h]
\vspace{-0.5cm}
\centering
\begin{center}
    \includegraphics[scale = 0.25]{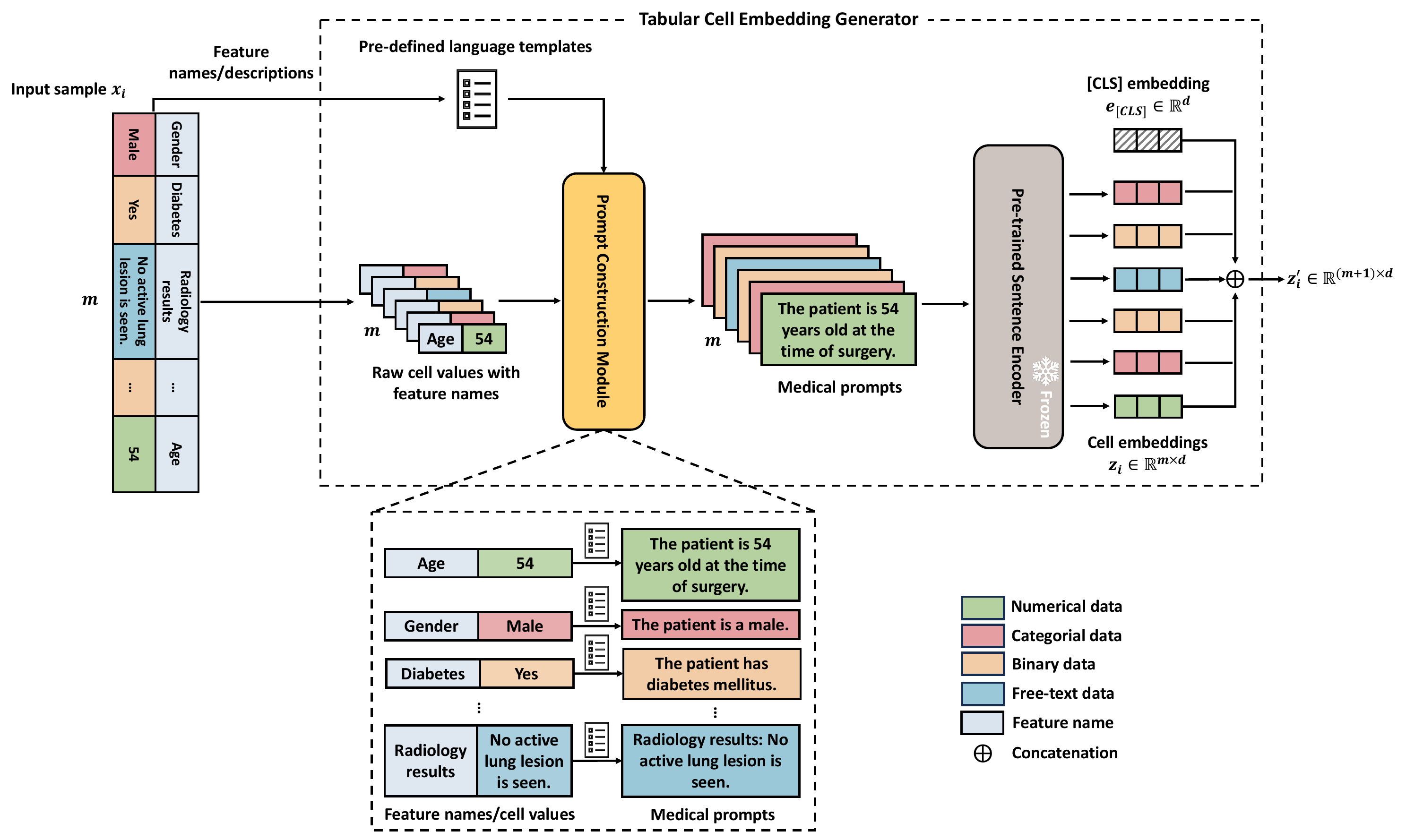}
\end{center}

\caption{Overview of tabular cell embedding generator. }
  \label{fig:tabular_cell_emb}
\vspace{-0.5cm}
\end{figure*}

The overview of tabular cell embedding generator is illustrated in Figure \ref{fig:tabular_cell_emb}, and it has two critical components: prompt construction module and pre-trained sentence encoder.

\subsubsection{Prompt Construction Module}

Inspired by the concept of prompt learning, we design a prompt construction module using medical language templates, which enables the pre-trained language model to better extract contextual information from medical tabular data and produce more comprehensive cell embeddings.
Instead of generating prompts with unfilled slots, our language templates transform raw cell values in tabular data into natural sentences, which the pre-trained sentence encoder can use to generate sentence embeddings as cell representations. In this way, the contextual information of raw cells can also be retrieved.

Let us denote $i$-th training sample in medical tabular dataset as $x_i = (c_{i,1}, c_{i,2}, \dots, c_{i,m})$, where $c_{i,j}$ is the cell value as string of $i$-th training sample under $j$-th column, and $m$ is the number of features (columns) in the dataset. In prompt construction module, we have pre-defined a set of medical language templates for each feature as $T = \{t_1, t_2, \dots, t_m\}$, in which $t_j$ denotes for the template for $j$-th feature to fill in. Therefore, we can obtain the medical prompts $s_{i,j}$ from cell $c_{i,j}$ by
$c_{i,j} \xrightarrow{t_j} s_{i,j}.$

For example, assuming our $k$-th feature is $weight$ in our dataset, the language template $t_k$ is constructed as "The weight of patient is $c_{i,k}$ kilograms". 
Different features are corresponded with different prompt templates. In general, after prompt construction module, medical prompts $q_i$ of $i$-th training sample, denoted by $q_i = (s_{i,1}, s_{i,2}, \dots, s_{i,m})$, have been generated for all cells to generate contextualized cell embeddings through the pre-trained sentence encoder.

\subsubsection{Pre-trained Sentence Encoder}

To further explore the textual information in medical tabular data, we adopt supervised SimCSE,\cite{gao-etal-2021-simcse} a RoBERTa-based \cite{liu2019roberta} framework with a contrastive learning objective, to advance the state-of-the-art in sentence representation for harmonized cell embeddings extraction.

Let us assume $s_{i,j} = (w_{i,j}^1, w_{i,j}^2, \dots, w_{i,j}^D)$ as the medical prompt (input sentence) of $i$-th training sample under $j$-th column, where $w_{i,j}^k$ is the $k$-th token in the sequence, and $D$ is the maximum number of tokens that the pre-trained model can take in. In pre-trained sentence encoder, $w_{i,j}^1$ is typically the special token $<$s$>$ marking the start of sequence. In cells where the length of tokens is shorter than $D$, the padding token $<$pad$>$ is appended at the end of the sentence to align with the maximum sentence length of $D$.
The output embedding $b_{i,j}^k$ of $k$-th token is obtained by:

\begin{equation}
    b_{i,j}^1, b_{i,j}^2, \dots, b_{i,j}^D = \mathrm{RoBERTa_{pre}} (w_{i,j}^1, w_{i,j}^2, \dots, w_{i,j}^D)
\end{equation}
Thereafter, we can obtain the cell embedding $z_{i,j}$ by:
\begin{equation}
    z_{i,j} = \mathrm{pooling}(b_{i,j}^1, b_{i,j}^2, \dots, b_{i,j}^D)
\end{equation}

$\mathrm{RoBERTa_{pre}(\cdot)}$ is denoted as the pre-trained RoBERTa model while pooling($\cdot$) is the mean pooling layer after token embeddings, and $z_{i,j} \in \mathbb{R}^{d}$ where $d$ is the embedding dimension of the pre-trained sentence encoder. Consequently, the total $m$ cell embeddings are $z_{i} \in \mathbb{R}^{m \times d}$. Following the existing research \cite{gorishniy2021revisiting,wang2022transtab}, [CLS] embedding $e_{[CLS]} \in \mathbb{R}^{d}$ has been concatenated with cell embeddings for patient embedding learning as $z'_i = e_{[CLS]} \oplus z_i \in \mathbb{R}^{(m+1) \times d}$.
As a result, the concatenated embeddings $z'_i$ are obtained for multimodality exploration in tabular transformer.
\subsection{Tabular Transformer and Prediction Head}
Since many existing approaches have demonstrated the effectiveness of transformer architecture in tabular domain, we adopt the classical transformer architecture \cite{vaswani2017attention} used in tabular domain \cite{gorishniy2021revisiting} to generate representative patient embedding, which removes the positional encoding at the inputs.

\begin{figure*}[!h]
\vspace{-0.5cm}
\centering
\begin{center}
    \includegraphics[scale = 0.32]{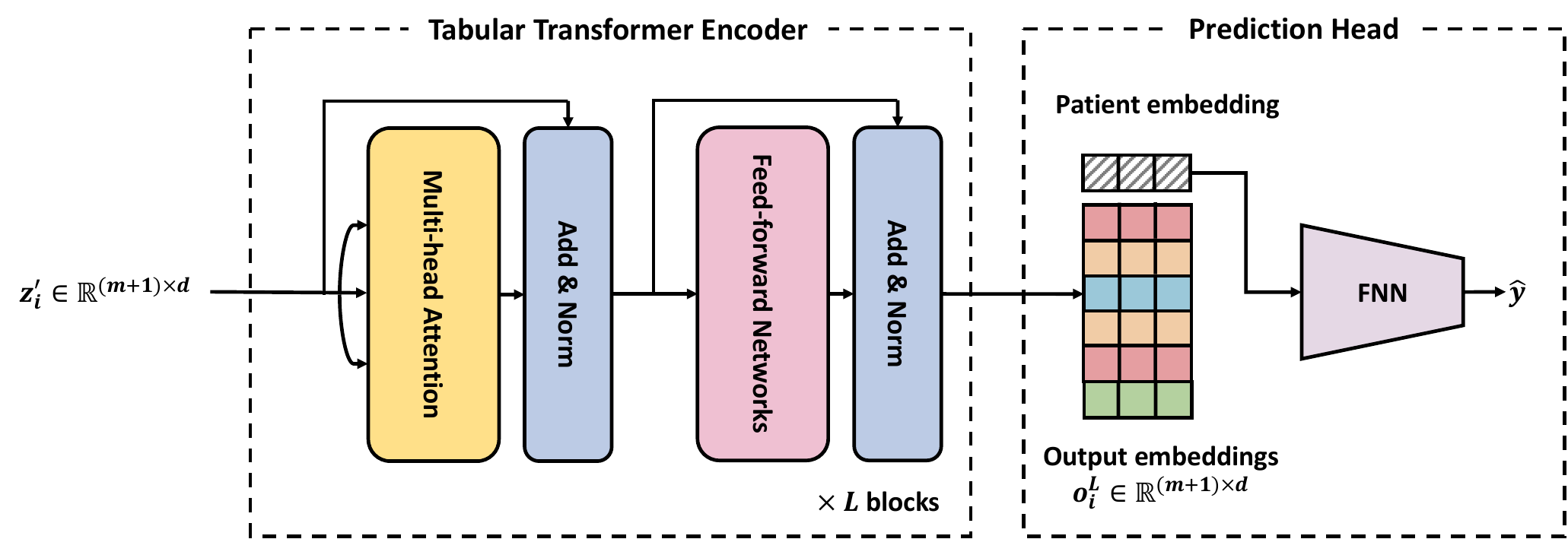}
\end{center}

\caption{Overview of tabular transformer and prediction head.}
  \label{fig:tabtran}
\vspace{-0.5cm}
\end{figure*}

As shown in Figure \ref{fig:tabtran}, given the output embeddings $z'_{i} \in \mathbb{R}^{(m+1) \times d}$ from the tabular cell encoder, the output embeddings of the entire tabular transformer architecture are $o_i^L \in \mathbb{R}^{(m+1) \times d}$. Then, in the prediction head, a feed-forward neural network with one hidden layer takes the [CLS] embedding $e'_{[CLS]} \in \mathbb{R}^d$ as the patient embedding for further prediction. 

\section{Experiments}\label{sec5}
\subsection{Datasets}
This study used two real-world datasets: a Asian dataset called PASA (Perioperative Anaesthesia Subject Area) and a US dataset called MIMIC-III (Medical Information Mart for Intensive Care III). These datasets contain 4 types of features: categorical, continuous, binary, and free-texts. We preprocessed and split the dataset into train, val, and test sets with a 3:1:1 ratio. Table \ref{tab:overall} describes the basic statistics of the datasets. 

\begin{table}[!h]
\vspace{-0.5cm}
  \caption{Descriptive statistics for PASA and MIMIC-III datasets}
  \centering
  \begin{subtable}[t]{0.45\textwidth}
    \centering
\caption{PASA}
\begin{tabular}{ll}
\hline
Description                        & Value \\ \hline
Total size                         & 71082 \\
No. of categorical features & 12    \\
No. of continuous features  & 26    \\
No. of binary features      & 32    \\
No. of free-text features        & 4     \\
Average surgical duration (hours)          & 2.50      \\ \hline
\end{tabular}
\label{tab: data_pasa}
  \end{subtable}%
  \hspace{0.05\textwidth}
  \begin{subtable}[t]{0.45\textwidth}
   \centering
\caption{MIMIC-III}
\begin{tabular}{ll}
\hline
Description                        & Value \\ \hline
Total size                         & 38468 \\
No. of categorical features & 3    \\
No. of continuous features  & 22    \\
No. of binary features      & 13    \\
No. of free-text features        & 4     \\
Mortality rate & 0.12 \\
Average LOS in ICU (days)          &   4.06     \\ 

\hline
\end{tabular}

\label{tab: data_mimic}
  \end{subtable}

  \label{tab:overall}
\vspace{-1cm}
\end{table}
 
\subsubsection{PASA dataset}

This surgical dataset was obtained retrospectively from Perioperative Anaesthesia Subject Area (PASA) of Singapore General Hospital between 2016 to 2020 \cite{chiew2020utilizing}. In the experiemtn, we aimed to estimate the surgical duration (SD) as a regression task.

\subsubsection{MIMIC-III dataset}

MIMIC-III is a large, publicly available datasets containing de-identified health records from patients in critical care units at Beth Israel Deaconess Medical Center (from US) between 2001 and 2012 \cite{johnson2016mimic}.
For this experiment, we have two prediction tasks: mortality and length of stay (LOS) in ICU, in which the former one is for binary classification task and the latter is for regression task.

\subsection{Baselines}
To better demonstrate the performance of the proposed model, we conduct comparison experiments on the both datasets with the following that have achieved great results in tabular prediction: Random Forest \cite{breiman2001random}, XGBoost \cite{chen2016xgboost}, MLP \cite{gorishniy2021revisiting}, ResNet \cite{gorishniy2021revisiting}, TabNet \cite{arik2021tabnet}, TabTransformer \cite{huang2020tabtransformer}, FT-Transformer \cite{gorishniy2021revisiting} and TransTab \cite{wang2022transtab}. 

\subsection{Implementation and Evaluation}

In proposed model, we used the supervised SimCSE \cite{gao-etal-2021-simcse} as the pre-trained sentence encoder. Each word token was mapped into a 768-dimensional embedding, and the entire encoder was frozen in the training process. The tabular transformer consisted of 6 basic transformer encoder layers, each with 6 heads in the attention layer. Throughout training, we used the Adam optimizer \cite{kingma2014adam} to update gradients with a learning rate of 1e-5 across all tasks. A mini-batch size was set to 256, and the maximum number of epochs was set to 100, with early stopping applied. 

We reported balanced accuracy (BACC) and the area under the ROC curve (AUROC) to effectively evaluate the classification task with the imbalanced dataset. For regression tasks, we used root mean squared error (RMSE) and mean absolute error (MAE) as evaluation metrics.
We repeated model training 5 times with different random seeds and reported the
average metrics, ensuring statistically stable results in this study.

\section{Results and Analyses}

\subsection{Model Performance}

\begin{table}[!h]
\vspace{-0.5cm}
\centering
\caption{The results of model comparisons on PASA and MIMIC-III datasets for three different tasks. The best and second-best results are in bold and underlined.}
\vspace{0.5cm}
\begin{tabular}{lllllll}
\hline
\multirow{2}{*}{Model} & \multicolumn{2}{l}{PASA(SD)} & \multicolumn{2}{l}{MIMIC(LOS)} & \multicolumn{2}{l}{MIMIC(Mortality)} \\ \cline{2-7} 
                       & RMSE$\downarrow$        & MAE$\downarrow$           & RMSE$\downarrow$          & MAE$\downarrow$           & BACC$\uparrow$             & AUROC$\uparrow$              \\ \hline
Random Forest          & 1.410        & 0.896        & 5.820         & 3.077         & 0.722            & 0.809            \\
XGBoost                & 1.364        & 0.845        & 5.731         & 3.003         & 0.761            & 0.846            \\
MLP                    & 1.376        & 0.851        & 5.771         & 3.027         & 0.754            & 0.836            \\
ResNet                 & 1.368        & 0.840        & 5.752         & 2.983         & 0.759            & 0.842            \\
TabNet                 & 1.423        & 0.876        & 5.846         & 3.081         & 0.748            & 0.826            \\
TabTransformer         & 1.353        & 0.848        & 5.754         & 3.053         & 0.754            & 0.835            \\
FT-Transformer         & 1.343        & 0.828        & \underline{5.722}         & 3.002         & 0.762            & \underline{0.848}            \\
TransTab               & 1.349        & 0.860        & 5.801         & 3.100         & 0.741            & 0.827            \\ \hline
Ours              & \textbf{1.197}       & \textbf{0.737}        & \textbf{5.692}         & \textbf{2.918}         & \textbf{0.774}            & \textbf{0.855}            \\
$\mathrm{\Delta}$ over best baseline &
%
%
\cellcolor{mycolor} 10.9\%       & \cellcolor{mycolor} 11.0\%        & \cellcolor{mycolor} 0.5\%         & \cellcolor{mycolor} 2.2\%         & \cellcolor{mycolor} 1.6\%            & \cellcolor{mycolor} 0.8\%            \\ \hdashline[2.5pt/5pt]
Ours (w/o free-texts)  \qquad            & \underline{1.340}        & \underline{0.825}        & 5.755         & \underline{2.964}         & \underline{0.765}            & 0.844            \\  \hline
\multicolumn{7}{l}{$\mathrm{\Delta}$: model performance improvements}                                                               

\end{tabular}
\label{tab:model_comparison}
\vspace{-0.5cm}
\end{table}

The results, presented in Table \ref{tab:model_comparison}, demonstrate the superior performance of our proposed framework across various tasks. Specifically, in the PASA(SD) task, our framework achieved the best performance among the current SOTA models.
To provide further insights, our framework demonstrated performance improvements, reducing RMSE/MAE by approximately 10.9\%/11.0\% in the PASA(SD) task and 0.5\%/2.2\% in the MIMIC(LOS) task compared to the best baseline. In the MIMIC(Mortality) task, our model exhibited a performance increase of about 1.6\%/0.8\% in BACC/AUROC over the best baseline. The consistent superiority of our model across tasks underscores its effectiveness. Additionally, we observed variations in model performance between datasets. The PASA dataset exhibited significantly enhanced model performance compared to the MIMIC-III dataset. This discrepancy might be attributed to the notably lower frequency of missing clinical free-texts in the PASA dataset. Specifically, there were about 15\% missing values in PASA dataset, whereas this figure rose to about 65\% in MIMIC data. This observation emphasizes the crucial role of free-text information in medical tabular prediction.

In addition, comparing our model's results with and without free-text columns showed that clinical free-text information consistently improved predictive performance.
Even without free-texts, our proposed model generally outperformed the best baselines, demonstrating its capability to leverage textual information in structured EHRs.

Within machine learning baselines, XGBoost continued to exhibit strong performance in the prediction of medical tabular data. In addition, FT-Transformer emerged as a compelling competitor to XGBoost in these tabular data models. Importantly, it is worth noting that, although TransTab incorporates textual information through the learning of word embeddings, these embeddings are not adequately learned with the constraints of relatively limited medical data when compared to the extensive data used in training language models. As a consequence, this inadequacy led to a degradation in predictive performance when compared to state-of-the-art transformer-based tabular models.

\section{Conclusion}

In this paper, we present a novel prompt-based tabular transformer framework, PTransformer, to model multimodalities in medical tabular data from an NLP perspective. 
PTransformer includes a tabular cell embedding generator, wherein a pre-trained sentence encoder and medical prompts collaborate to generate contextualized cell embeddings in a harmonized language space; 
It also incorporates a tabular transformer to generate informative patient embeddings for prediction.
Experiments on two large medical datasets show that PTransformer consistently outperforms other SOTA baselines 
and highlight the benefits of prompt-based learning with transformer-based framework to address multimodality modeling in medical tabular data.
In future work, we plan to explore soft prompts that can be optimized during training for specific tasks.

\begin{credits}
\subsubsection{\ackname} This research is supported by the National Research Foundation Singapore under its AI Singapore Programme grant number AISG-GC-2019-001-2A. This research is also supported by A*STAR, CISCO Systems (USA) Pte. Ltd and National University of Singapore under its Cisco-NUS Accelerated Digital Economy Corporate Laboratory (Award I21001E0002).

\subsubsection{\discintname}
The authors have no competing interests to declare.
\end{credits}
%
%
%
\bibliographystyle{splncs04}
\bibliography{sn-bibliography}

\begin{thebibliography}{10}
\providecommand{\url}[1]{\texttt{#1}}
\providecommand{\urlprefix}{URL }
\providecommand{\doi}[1]{https://doi.org/#1}

\bibitem{arik2021tabnet}
Arik, S.{\"O}., Pfister, T.: Tabnet: Attentive interpretable tabular learning. In: Proceedings of the AAAI conference on artificial intelligence. vol.~35, pp. 6679--6687 (2021)

\bibitem{breiman2001random}
Breiman, L.: Random forests. Machine learning  \textbf{45}(1),  5--32 (2001)

\bibitem{chen2016xgboost}
Chen, T., Guestrin, C.: Xgboost: A scalable tree boosting system. In: Proceedings of the 22nd acm sigkdd international conference on knowledge discovery and data mining. pp. 785--794 (2016)

\bibitem{chen2021automatic}
Chen, X., Huang, L., Liu, W., Shih, P.C., Bao, J.: Automatic surgery duration prediction using artificial neural networks. In: The 5th International Conference on Computer Science and Application Engineering. pp.~1--6 (2021)

\bibitem{chiew2020utilizing}
Chiew, C.J., Liu, N., Wong, T.H., Sim, Y.E., Abdullah, H.R.: Utilizing machine learning methods for preoperative prediction of postsurgical mortality and intensive care unit admission. Annals of surgery  \textbf{272}(6), ~1133 (2020)

\bibitem{gao-etal-2021-simcse}
Gao, T., Yao, X., Chen, D.: Sim{CSE}: Simple contrastive learning of sentence embeddings. In: Proceedings of the 2021 Conference on Empirical Methods in Natural Language Processing. pp. 6894--6910 (Nov 2021). \doi{10.18653/v1/2021.emnlp-main.552}

\bibitem{gao2022prediction}
Gao, W., Wang, J., Zhou, L., Luo, Q., Lao, Y., Lyu, H., Guo, S.: Prediction of acute kidney injury in icu with gradient boosting decision tree algorithms. Computers in biology and medicine  \textbf{140},  105097 (2022)

\bibitem{george2021deep}
George, N., Moseley, E., Eber, R., Siu, J., Samuel, M., Yam, J., Huang, K., Celi, L.A., Lindvall, C.: Deep learning to predict long-term mortality in patients requiring 7 days of mechanical ventilation. PloS one  \textbf{16}(6),  e0253443 (2021)

\bibitem{gorishniy2021revisiting}
Gorishniy, Y., Rubachev, I., Khrulkov, V., Babenko, A.: Revisiting deep learning models for tabular data. Advances in Neural Information Processing Systems  \textbf{34},  18932--18943 (2021)

\bibitem{grinsztajn2022tree}
Grinsztajn, L., Oyallon, E., Varoquaux, G.: Why do tree-based models still outperform deep learning on typical tabular data? Advances in Neural Information Processing Systems  \textbf{35},  507--520 (2022)

\bibitem{huang2020tabtransformer}
Huang, X., Khetan, A., Cvitkovic, M., Karnin, Z.: Tabtransformer: Tabular data modeling using contextual embeddings. arXiv preprint arXiv:2012.06678  (2020)

\bibitem{johnson2016mimic}
Johnson, A.E., Pollard, T.J., Shen, L., Lehman, L.w.H., Feng, M., Ghassemi, M., Moody, B., Szolovits, P., Anthony~Celi, L., Mark, R.G.: Mimic-iii, a freely accessible critical care database. Scientific data  \textbf{3}(1), ~1--9 (2016)

\bibitem{kingma2014adam}
Kingma, D.P., Ba, J.: Adam: A method for stochastic optimization. In: The 3rd International Conference on Learning Representations, {ICLR}  (2015)

\bibitem{liu2019roberta}
Liu, Y., Ott, M., Goyal, N., Du, J., Joshi, M., Chen, D., Levy, O., Lewis, M., Zettlemoyer, L., Stoyanov, V.: Roberta: A robustly optimized bert pretraining approach. arXiv preprint arXiv:1907.11692  (2019)

\bibitem{nistal2022developing}
Nistal-Nu{\~n}o, B.: Developing machine learning models for prediction of mortality in the medical intensive care unit. Computer Methods and Programs in Biomedicine  \textbf{216},  106663 (2022)

\bibitem{przystalski2023medical}
Przystalski, K., Thanki, R.M.: Medical tabular data. In: Explainable Machine Learning in Medicine, pp. 17--36. Springer (2023)

\bibitem{vaswani2017attention}
Vaswani, A., Shazeer, N., Parmar, N., Uszkoreit, J., Jones, L., Gomez, A.N., Kaiser, {\L}., Polosukhin, I.: Attention is all you need. Advances in neural information processing systems  \textbf{30} (2017)

\bibitem{wang2022transtab}
Wang, Z., Sun, J.: Transtab: Learning transferable tabular transformers across tables. Advances in Neural Information Processing Systems  \textbf{35},  2902--2915 (2022)

\end{thebibliography}

\end{document}